# A Review Paper: Noise Models in Digital Image Processing

Ajay Kumar Boyat[1] and Brijendra Kumar Joshi[2]

[1]Research Scholar, Department of Electronics Telecomm and Computer Engineering, Military College of Tele Communication Engineering, Military Head Quartar of War (MHOW), Ministry of Defence, Govt. of India, India

[2]Professor, Department of Electronics Telecomm and Computer Engineering, Military College of Tele Communication Engineering, Military Head Quartar of War (MHOW), Ministry of Defence, Govt. of India, India

## ABSTRACT

Noise is always presents in digital images during image acquisition, coding, transmission, and processing steps. Noise is very difficult to remove it from the digital images without the prior knowledge of noise model. That is why, review of noise models are essential in the study of image denoising techniques. In this paper, we express a brief overview of various noise models. These noise models can be selected by analysis of their origin. In this way, we present a complete and quantitative analysis of noise models available in digital images.

## 1. INTRODUCTION

Many Practical developments, of considerable interest in the field of image denoising, need continuous and uniform review of relevant noise theory. Behalf of this, many researchers have addressed literature survey of given practical as well as theoretical aspects.

Although all literatures address the noise in imaging system usually presents during image acquisition, coding, transmission, and processing steps. This noise appearance disturbs the original information in voice, image and video signal. In this sense some questions arises in researches mind, how much original signal is corrupted?, how we can reconstruct the signal?, which noise model is associated in the noisy image.

However time to time we have to need the reinforcement learning of theoretical and practical ideas of entilt noises present in digital images. Here, we are trying to present the solution of all these problems through the review of noise models.

In this paper, the literature survey is based on statistical concepts of noise theory. We start with noise and the roll of noise in image distortion. Noise is random signal. It is used to destroy most of the part of image information. Image distortion is most pleasance problems in image processing. Image distorted due to various types of noise such as Gaussian noise, Poisson noise, Speckle noise, Salt and Pepper noise and many more are fundamental noise types in case of digital images. These noises may be came from a noise sources present in the vicinity of image capturing devices, faulty memory location or may be introduced due to imperfection/inaccuracy in the image capturing devices like cameras, misaligned lenses, weak focal length, scattering and other adverse conditions may be present in the atmosphere. This makes careful and in-depth study of noise and noise models are essential ingredient in image denoising. This leads to selection of proper noise model for image denoising systems [1-3].

## 2. NOISE MODELS

Noise tells unwanted information in digital images. Noise produces undesirable effects such as artifacts, unrealistic edges, unseen lines, corners, blurred objects and disturbs background scenes. To reduce these undesirable effects, prior learning of noise models is essential for further processing. Digital noise may arise from various kinds of sources such as Charge Coupled Device (CCD) and Complementary Metal Oxide Semiconductor (CMOS) sensors. In some sense, points spreading function (PSF) and modulation transfer function (MTF) have been used for timely, complete and quantitative analysis of noise models. Probability density function (PDF) or Histogram is also used to design and characterize the noise models. Here we will discuss few noise models, their types and categories in digital images [4].

### 2.1 Gaussian Noise Model

It is also called as electronic noise because it arises in amplifiers or detectors. Gaussian noise caused by natural sources such as thermal vibration of atoms and discrete nature of radiation of warm objects [5].

Gaussian noise generally disturbs the gray values in digital images. That is why Gaussian noise model essentially designed and characteristics by its PDF or normalizes histogram with respect to gray value. This is given as

$$P(g) = \sqrt{\frac{1}{2\pi\sigma^2}} e^{-\frac{(g-\mu)^2}{2\sigma^2}} \qquad (1)$$

Where g = gray value, $\sigma$ = standard deviation and $\mu$ = mean. Generally Gaussian noise mathematical model represents the correct approximation of real world scenarios. In this noise model, the mean value is zero, variance is 0.1 and 256 gray levels in terms of its PDF, which is shown in Fig. 1.

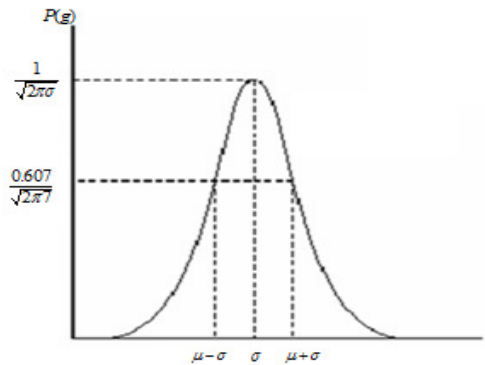

Figure 1 PDF of Gaussian noise

Due to this equal randomness the normalized Gaussian noise curve look like in bell shaped. The PDF of this noise model shows that 70% to 90% noisy pixel values of degraded image in between $\mu - \sigma$ and $\mu + \sigma$. The shape of normalized histogram is almost same in spectral domain.

## 2.2 White Noise

Noise is essentially identified by the noise power. Noise power spectrum is constant in white noise. This noise power is equivalent to power spectral density function. The statement "Gaussian noise is often white noise" is incorrect [4].

However neither Gaussian property implies the white sense. The range of total noise power is -∞ to +∞ available in white noise in frequency domain. That means ideally noise power is infinite in white noise. This fact is fully true because the light emits from the sun has all the frequency components.

In white noise, correlation is not possible because of every pixel values are different from their neighbours. That is why autocorrelation is zero. So that image pixel values are normally disturb positively due to white noise.

## 2.3 Brownian Noise (Fractal Noise)

Colored noise has many names such as Brownian noise or pink noise or flicker noise or 1/f noise. In Brownian noise, power spectral density is proportional to square of frequency over an octave i.e., its power falls on ¼ th part (6 dB per octave). Brownian noise caused by Brownian motion. Brownian motion seen due to the random movement of suspended particles in fluid. Brownian noise can also be generated from white noise, which is shown in Fig. 2.

However this noise follows non stationary stochastic process. This process follows normal distribution. Statistically fractional Brownian noise is referred to as fractal noise. Fractal noise is caused by natural process. It is different from Gaussian process [8-12].

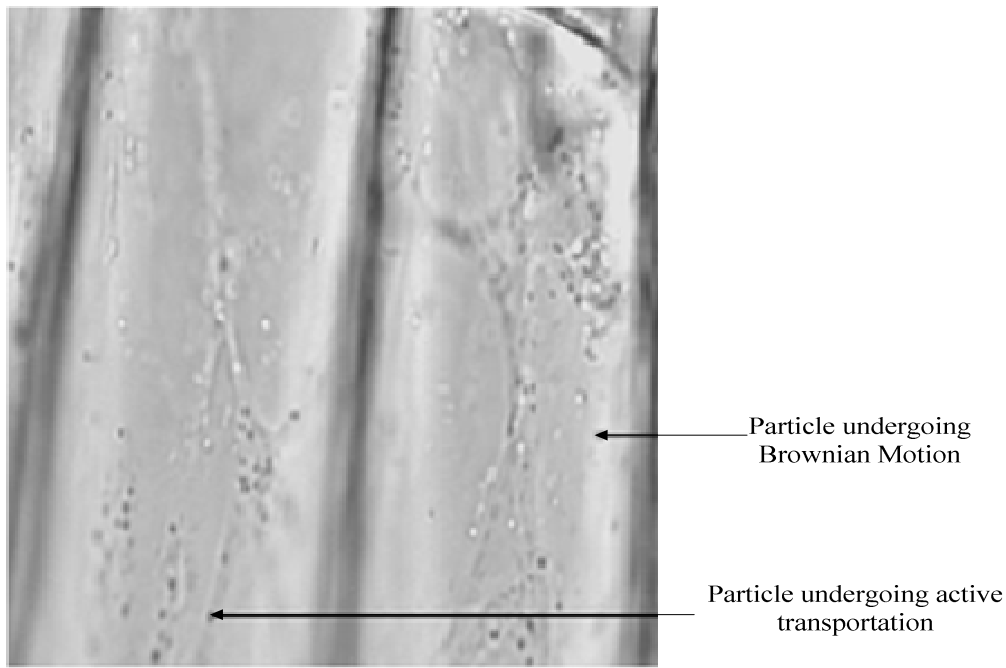

Figure 2 Visualized vesicles in onion cells (x20) form Brownian motion [10]

Although power spectrum of fractal noise, decaying continuously due to increase in frequency. Fractal noise is almost singular everywhere. A fractional Brownian motion is mathematically represents as a zero mean Gaussian process ($B_H$) which is showing in equation (2) and equation (3), respectively [6].

$$B_H(0) = 0 \qquad\qquad\qquad\qquad\qquad\qquad (2)$$

and expected value of fractional Brownian motion is

$$E\{\left|B_H(t) - B_H(t-\Delta)\right|^2\} = \sigma^2 \left|\Delta\right|^{2H} \qquad\qquad (3)$$

## 2.4 Impulse Valued Noise (Salt and Pepper Noise)

This is also called data drop noise because statistically its drop the original data values. This noise is also referred as salt and pepper noise. However the image is not fully corrupted by salt and pepper noise instead of some pixel values are changed in the image. Although in noisy image, there is a possibilities of some neighbours does not changed [13-14].

This noise is seen in data transmission. Image pixel values are replaced by corrupted pixel values either maximum 'or' minimum pixel value i.e., 255 'or' 0 respectively, if number of bits are 8 for transmission.

Let us consider 3x3 image matrices which are shown in the Fig. 3. Suppose the central value of matrices is corrupted by Pepper noise. Therefore, this central value i.e., 212 is given in Fig. 3 is replaced by value zero.

In this connection, we can say that, this noise is inserted dead pixels either dark or bright. So in a salt and pepper noise, progressively dark pixel values are present in bright region and vice versa [15].

| 254 | 207 | 210 |
|-----|-----|-----|
| 97  | 212 | 32  |
| 62  | 106 | 20  |

| 254 | 207 | 210 |
|-----|-----|-----|
| 97  | 0   | 32  |
| 62  | 106 | 20  |

Figure 3 The central pixel value is corrupted by Pepper noise

Inserted dead pixel in the picture is due to errors in analog to digital conversion and errors in bit transmission. The percentagewise estimation of noisy pixels, directly determine from pixel metrics. The PDF of this noise is shown in the Fig. 4.

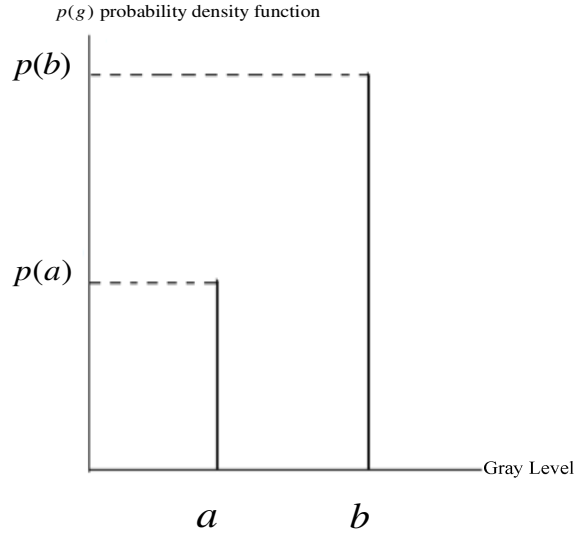

Figure 4. The PDF of Salt and Pepper noise

$$P(g) = \begin{cases} Pa \text{ for } g = a \\ Pb \text{ for } g = b \\ 0 \text{ otherwise} \end{cases} \qquad (4)$$

Fig. 4 shows the PDF of Salt and Pepper noise, if mean is zero and variance is 0.05. Here we will meet two spike one is for bright region (where gray level is less) called 'region a' and another one is dark region (where gray level is large) called 'region b', we have clearly seen here the PDF values are minimum and maximum in 'region a' and 'region b', respectively [16]. Salt and Pepper noise generally corrupted the digital image by malfunctioning of pixel elements in camera sensors, faluty memory space in storage, errors in digitization process and many more.

## 2.5 Periodic Noise

This noise is generated from electronics interferences, especially in power signal during image acquisition. This noise has special characteristics like spatially dependent and sinusoidal in nature at multiples of specific frequency. It's appears in form of conjugate spots in frequency domain. It can be conveniently removed by using a narrow band reject filter or notch filter.

## 2.6 Quantization noise

Quantization noise appearance is inherent in amplitude quantization process. It is generally presents due to analog data converted into digital data. In this noise model, the signal to noise ratio (SNR) is limited by minimum and maximum pixel value, $P_{min}$ and $P_{max}$ respectively.

The SNR is given as

$$SNR_{dB} = 20 \log_{10}(P_{max} - P_{min}) / \sigma_n \qquad (5)$$

Where $\sigma_n$ = Standard deviation of noise, when input is full amplitude sine wave SNR becomes

$$SNR = 6n + 1.76 \text{ dB} \qquad (6)$$

Where $n$ is number of bits. Quantization noise obeys the uniform distribution. That is why it is referred as uniform noise. Its PDF is shown in Fig. 5.

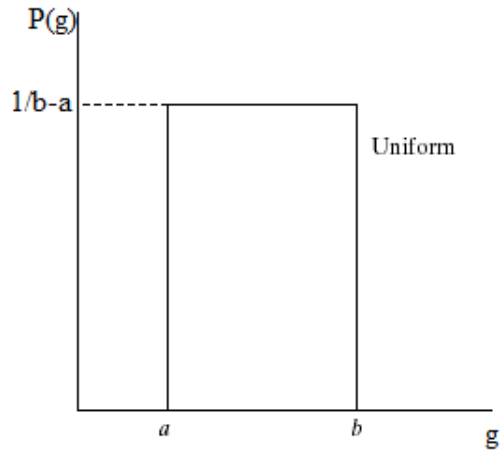

Figure 5 Uniform noise

$$P(g) = \begin{cases} \dfrac{1}{b-a} & \text{if } a \leq g \leq b \\ 0 & \text{otherwise} \end{cases} \qquad (7)$$

and their mean $\mu = \dfrac{a+b}{2}$ and variance $\sigma^2 = \dfrac{(b-a)^2}{12}$

## 2.7 Speckle Noise

This noise is multiplicative noise. Their appearance is seen in coherent imaging system such as laser, radar and acoustics etc,. Speckle noise can exist similar in an image as Gaussian noise. Its probability density function follows gamma distribution, which is shown in Fig. 6 and given as in equation (8) [17-19].

$$F(g) = \dfrac{g^{\alpha-1} e^{\frac{-g}{a}}}{\alpha-1! \, a^{\alpha}} \qquad (8)$$

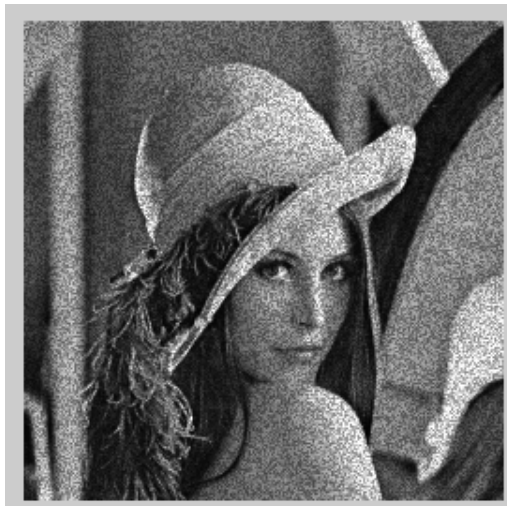

Figure 6 Lena image [20] of Speckle noise with variance 0.04

## 2.8 Photon Noise (Poisson Noise)

The appearance of this noise is seen due to the statistical nature of electromagnetic waves such as x-rays, visible lights and gamma rays. The x-ray and gamma ray sources emitted number of photons per unit time. These rays are injected in patient's body from its source, in medical x rays and gamma rays imaging systems. These sources are having random fluctuation of photons. Result gathered image has spatial and temporal randomness. This noise is also called as quantum (photon) noise or shot noise. This noise obeys the Poisson distribution and is given as

$$P(f_{(pi)} = k) = \frac{\lambda^k{}_i e^{-\lambda}}{k!} \qquad (9)$$

## 2.9 Poisson-Gaussian Noise

In this section the proposed model work for removing of Poisson-Gaussian noise that is arose in Magnetic Resonance Imaging (MRI). Poisson-Gaussian noise is shown in Fig. 7. The paper introduces the two most fantastic noise models, jointly called as Poisson-Gaussian noise model. These two noise models are specified the quality of MRI recipient signal in terms of visual appearances and strength [21]. Despite from the highest quality MRI processing, above model describes the set of parameters of the Poisson-Gaussian noise corrupted test image. The Poisson-Gaussian noise model can be illustrated in the following manner.

$$Z(j,k) = \alpha * P_\alpha(j,k) + N_\alpha(j,k) \qquad (10)$$

Where, the model has carried Poisson-distribution $(p_\alpha)$ follows with Gaussian-distribution $(n_\alpha)$. Poisson-distribution estimating using mean $(\mu_{\alpha 1})$ at given level $\alpha > 0$ and Gaussian-distribution counted at given level $\alpha > 0$ using zero mean $(\mu_{\alpha 2})$ and variance $\sigma^2_{\alpha 2}$. To evaluate $\mu_{\alpha 1}$, noisy Poisson image was quantitatively added to underlying original image. After all to get a noisy image $Z(j,k)$ from the Poisson-Gaussian model is based on [22-23].

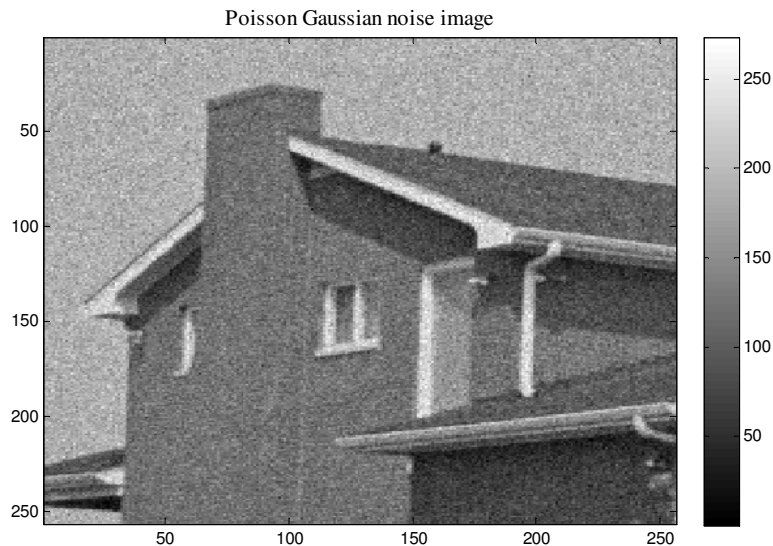

Figure 7 Poisson-Gaussian Noise House Image [20]

## 2.10 Structured Noise

Structured noise are periodic, stationary or non stationary and aperiodic in nature. If this noise is stationary, it has fixed amplitude, frequency and phase. Structured noise caused by interferences among electronic components [24]. Noise presents in communication channel are in two parts, unstructured noise (u) and structured noise (s). structured noise is also called low rank noise. In a signal processing, it is more advantagable (more realistic) to considering noise model in a lower dimensionality space.

Further, this model is mapped into full rank measurement space in physical system. So we can conclude that in the measurement space, resulting noise has low rank and exhibits structure dependent on physical system.

Structured noise model is showing in equation (11) and equation (12), respectively [25].

$$y_{(n)} = x_{(n,m)} + v_{(n)} \tag{11}$$

$$y_{(n)} = H_{(n,m)} * \theta_{(m)} + S_{(n,t)} * \phi_{(t)} + v_{(n)} \tag{12}$$

Where, $n =$ rows, $m =$ columns, $y =$ received image, $H =$ Transfer function of linear system, $S =$ Subspace, $t =$ rank in subspace, $\phi =$ underlying process exciting the linear system (S), $\theta =$ signal parameter sets initial conditions or excites, linear system H is used to produce original signal $x$ in terms of $n$ vector random noise $\left( v_{(n)} \right)$.

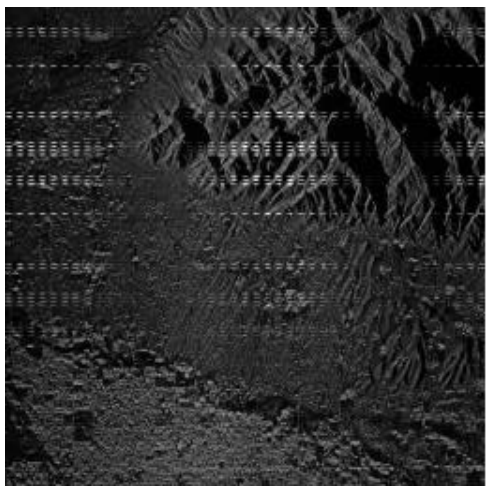

Figure 8 Structured Noise (when noise is periodic and non stationary) [25]

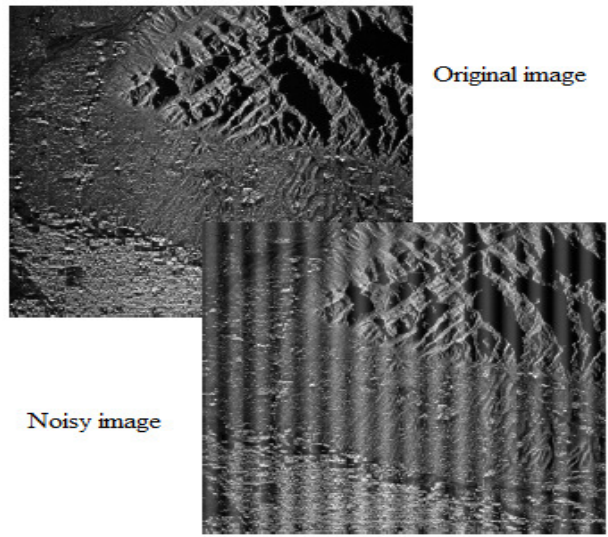

Figure 9 Structured Noise [25]

## 2.11 Gamma Noise

Gamma noise is generally seen in the laser based images. It obeys the Gamma distribution. Which is shown in the Fig. 10 and given as [26-27]

$$P(g) = \begin{cases} \dfrac{a^b g^{b-1} e^{-ag}}{(b-1)!} & \text{for } g \ge 0 \\ 0 & \text{for } g < 0 \end{cases} \qquad (13)$$

Where mean $\mu = \dfrac{b}{a}$ and variance $\sigma^2 = \dfrac{b}{a^2}$ are given as, respectively.

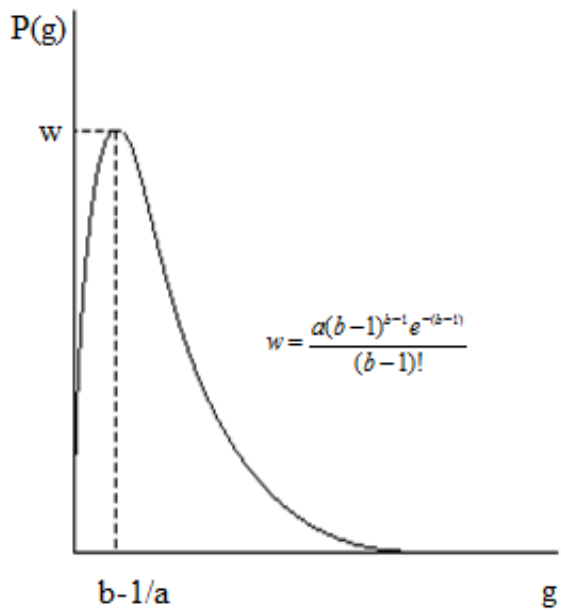

Figure 10 Gamma distribution

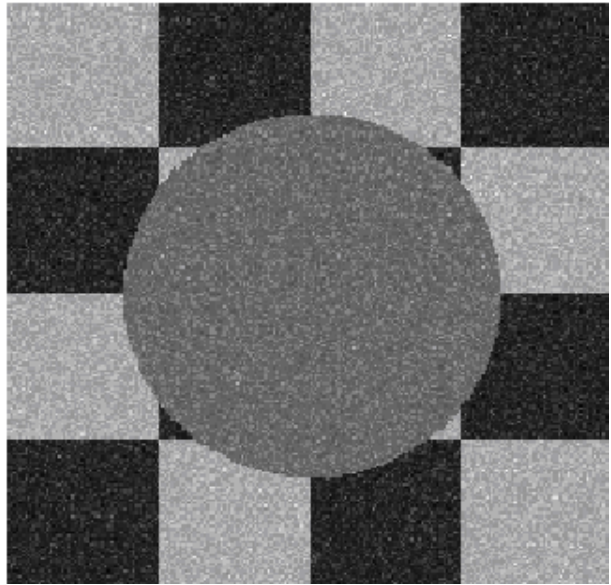

Figure 11 Gamma Noise [26]

## 2.12 Rayleigh noise

Rayleigh noise presents in radar range images. In Rayleigh noise, probability density function is given as [26].

$$P(g) = \begin{cases} \dfrac{2}{b}(g-a)e^{\frac{-(g-a)^2}{b}} & \text{for } g \geq a \\ 0 & \text{for } g < a \end{cases} \tag{14}$$

Where mean $\mu = a + \sqrt{\dfrac{\pi b}{4}}$ and variance $\sigma^2 = \dfrac{b(4-\pi)}{4}$ are given as, respectively.

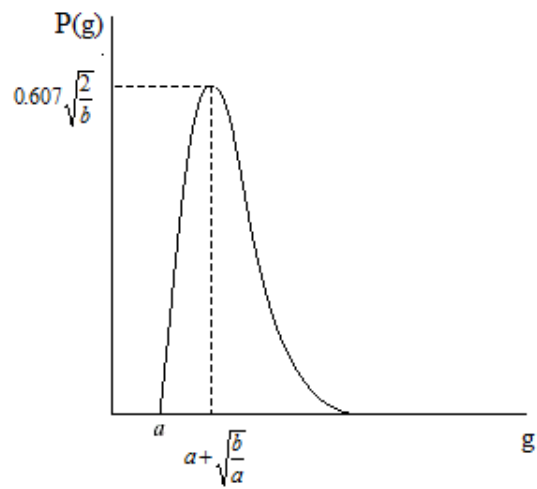

Figure 12 Rayleigh distribution

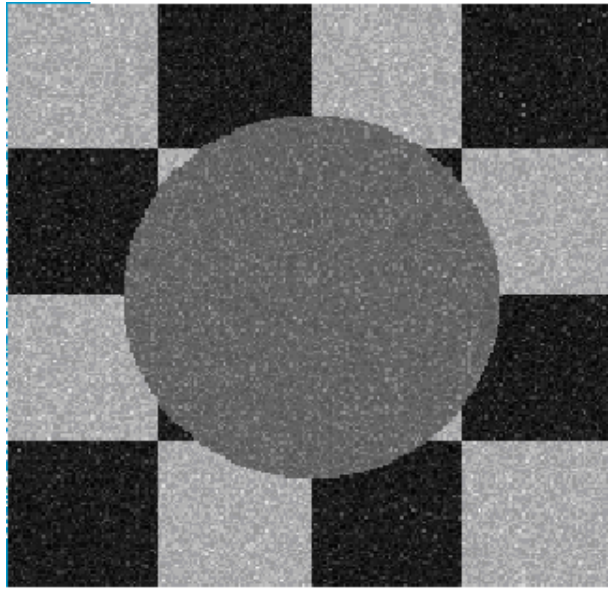

Figure 13 Rayleigh Noise [26]

## 3. CONCLUSIONS

During image acquisition and transmission, noise is seen in images. This is characterised by noise model. So study of noise model is very important part in image processing. On the other hand, Image denoising is necessary action in image processing operation. Without the prior knowledge of noise model we cannot elaborate and perform denoising actions.

Hence, here we have reviewed and presented various noise models available in digital images. We addressed that noise models can be identified with the help of their origin. Noise models also designed by probability density function using mean, variance and mainly gray levels in digital images. We hope this work will provide as a susceptible material for researchers and of course for freshers in the image processing field.

## AUTHORS

Ajay Boyat is a Research scholar in Electronics & Telecommunication and Computer Engineering Department at Military College of Telecommunication Engineering (MCTE), Military Head Quarter of War (MHOW), M.P., India. He has completed B.E. in Electronics and Communication Engineering from MITS, Gwalior in 2005, M.E. in Electronics and Telecommunication Engineering from SGSITS, Indore in 2008 and Currently Pursuing Ph.D. in Electronics and Telecommunication Engineering from Devi 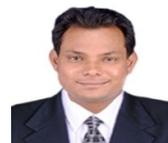
Ahiliya University, Indore. His research interests are Image processing, Adaptive and Non-Linear Filtering, Signal Processing, Mobile adhoc and Wireless Communication networks. He has number of research papers to his credit and has supervised three Post Graduate scholar and currently supervising three Post Graduate scholars. He has 09 years of teaching experience. He is Research and Project Cordinator in Department of Electronics & Communication Engineering at Medi-Caps Institute of Technology and Management, Indore (MP), India. He is a Global member of Internet Society. Recently he won the Best paper award Rank 1 in track 5 in 2014 IEEE ICCIC from IEEE PODHIGAI and IEEE-SIPCICOM.

Dr. Brijendra Kumar Joshi is a Professor in Electronics & Telecommunication and Computer Engineering Department at Military College of Telecommunication Engineering (MCTE), Military Head Quarter of War (MHOW), M.P., and India. Dr. Joshi, Fellow IETE India and IE(I). He is Life time Senior Member of CSI and Life Member of ISTE.  He is Member of Cryptology Research Society of India (CRSI) and Member of IEEE, IEEE Communication Society and ACM (USA). He has completed 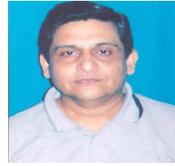 B.E. in Electronics and Telecommunication Engineering from Govt. Engineering. College, Jabalpur, M.E. in Computer Science and Engineering from Indian Institute of Science, Bangalore, Ph.D. in Electronics and Telecommunication Engineering from Rani Durgavati University, Jabalpur and M. Tech. in Digital Communication from MANIT, Bhopal. His research interests are Programming Languages, Compiler Design, Digital Communications, Mobile adhoc and Wireless Sensor Networks, Signal Processing, Parallel Computing and Software Engineering. He has number of research papers to his credit and has supervised three PhD scholars and currently supervising eight PhD scholars. He has published two books on Data Structures and Algorithms with Tata McGraw-Hill. He has 30 years of teaching experience.